\documentclass[ba]{imsart}
\pubyear{0000}
\volume{00}
\issue{0}
\doi{0000}
\firstpage{1}
\lastpage{1}

\usepackage{amsthm}
\usepackage{amsmath}
\usepackage{natbib}
\usepackage[colorlinks,citecolor=blue,urlcolor=blue,filecolor=blue,backref=page]{hyperref}
\usepackage{graphicx}

\usepackage{amsfonts}
\usepackage{amssymb}
\usepackage[ruled,vlined, linesnumbered]{algorithm2e}

\newtheorem{proposition}{Proposition}

\begin{document}

\begin{frontmatter}
\title{Constrained Bayesian Optimization with Noisy Experiments}
\runtitle{Constrained Bayesian Optimization with Noisy Experiments}

\begin{aug}
\author{\fnms{Benjamin} \snm{Letham}\thanksref{addr1}\ead[label=e1]{bletham@fb.com}},
\author{\fnms{Brian} \snm{Karrer}\thanksref{addr2}\ead[label=e2]{briankarrer@fb.com}},
\author{\fnms{Guilherme} \snm{Ottoni}\thanksref{addr3}\ead[label=e3]{ottoni@fb.com}},
\and
\author{\fnms{Eytan} \snm{Bakshy}\thanksref{addr4}\ead[label=e4]{ebakshy@fb.com}}

\runauthor{B. Letham et al.}

\address[addr1]{Facebook, Menlo Park, California, USA
\printead{e1}}
\address[addr2]{Facebook, Menlo Park, California, USA
\printead{e2}}
\address[addr3]{Facebook, Menlo Park, California, USA
\printead{e3}}
\address[addr4]{Facebook, Menlo Park, California, USA
\printead{e4}}
\end{aug}

\begin{abstract} 
Randomized experiments are the gold standard for evaluating the effects of changes to real-world systems. Data in these tests may be difficult to collect and outcomes may have high variance, resulting in potentially large measurement error. Bayesian optimization is a promising technique for efficiently optimizing multiple continuous parameters, but existing approaches degrade in performance when the noise level is high, limiting its applicability to many randomized  experiments. We derive an expression for expected improvement under greedy batch optimization with noisy observations and noisy constraints, and develop a quasi-Monte Carlo approximation that allows it to be efficiently optimized. Simulations with synthetic functions show that optimization performance on noisy, constrained problems outperforms existing methods. We further demonstrate the effectiveness of the method with two real-world experiments conducted at Facebook: optimizing a ranking system, and optimizing server compiler flags.
\end{abstract} 


\end{frontmatter}

\section{Introduction}
Many policies and systems found in Internet services, medicine,  economics, and other settings have continuous parameters that affect outcomes of interest that can only be measured via randomized experiments.
These design parameters often have complex interactions that make it impossible to know \textit{a priori} how they should be set to achieve the best outcome. Randomized experiments, commonly referred to as A/B tests in the Internet industry, provide a mechanism for directly measuring the outcomes of any given set of parameters, but they are typically time consuming and utilize a limited resource of available samples.  As a result, many systems used in practice involve various constants that have been chosen with a limited amount of manual tuning.

Bayesian optimization is a powerful tool for solving black-box global optimization problems with computationally expensive function evaluations \citep{jones98}. Most commonly, this process begins by evaluating a small number of randomly selected function values, and fitting a Gaussian process (GP) regression model to the results. The GP posterior provides an estimate of the function value at each point, as well as the uncertainty in that estimate. We then choose a new point at which to evaluate the function by balancing exploration (high uncertainty) and exploitation (best estimated function value). This is done by optimizing an acquisition function, which encodes the value of potential points in the optimization and defines the balance between exploration and exploitation. A common choice for the acquisition function is \textit{expected improvement} (EI), which measures the expected value of the improvement at each point over the best observed point. Optimization then continues sequentially, at each iteration updating the model to include all past observations.

Bayesian optimization has recently become an important tool for optimizing machine learning hyperparameters \citep{snoek12}, where in each iteration a machine learning model is fit to data and prediction quality is observed. Our work is motivated by a need to develop robust algorithms for optimizing via randomized experiments. There are three aspects of A/B tests that differ from the usual hyperparameter optimization paradigm. First, there are typically high noise levels when measuring performance of systems. Extensions of Bayesian optimization to handle noisy observations use heuristics to simplify the acquisition function that can perform poorly with high noise levels. Second, there are almost always trade-offs involved in optimizing real systems: improving the quality of images may result in increased data usage; increasing cache sizes may improve the speed of a mobile application, but decrease reliability on some devices. Practitioners have stressed the importance of considering multiple outcomes \citep{deng2016data}, and noisy constraints must be incorporated into the optimization. Finally, it is often straightforward to run multiple A/B tests in parallel, with limited wall time in which to complete the optimization. Methods for batch optimization have been developed in the noiseless case; here we unify the approach for handling noise and batches.

This work is related to policy optimization \citep{athey17}, which seeks to learn an optimal mapping from context to action. When the action space is discrete this is the classic contextual bandit problem \citep{dudik2014doubly}, but with a continuous action space it can be solved using Bayesian optimization. For example, there are many continuous parameters involved in encoding a video for upload and the most appropriate settings depend on the Internet connection speed of the device. We can use Bayesian optimization to learn a policy that maps connection speed to encoding parameters by including connection speed in the model feature space. Related policy optimization problems can be found in medicine \citep{zhao12} and reinforcement learning \citep{wilson2014bo4rl, marco17}.

Most work in Bayesian optimization does not handle noise in a Bayesian way. We derive a Bayesian expected improvement under noisy observations and noisy constraints that avoids simplifying heuristics by directly integrating over the posterior of the acquisition function. We show that this can be efficiently optimized via a quasi-Monte Carlo approximation. We have used this method at Facebook to run dozens of optimizations via randomized experiments, and here demonstrate the applicability of Bayesian optimization to A/B testing with two such examples: experiments to tune a ranking system, and optimizing server compiler settings.

\section{Prior work on expected improvement}\label{sec:prior}
The EI acquisition function was introduced by \citet{jones98} for efficient optimization of computationally expensive black-box functions. They considered an unconstrained problem $\min_{\mathbf{x}} f(\mathbf{x})$ with noiseless function evaluations. Given data $\mathcal{D}_f = \{\mathbf{x}_i, f(\mathbf{x}_i)\}_{i=1}^n$, we first use GP regression to estimate $f$. Let $g(\mathbf{x} | \mathcal{D}_f)$ be the GP posterior at $\mathbf{x}$ and $f^* = \min_i f(\mathbf{x}_i)$ the current best observation. The EI of a candidate $\mathbf{x}$ is the expectation of its improvement over $f^*$:
\begin{equation*}
\alpha_{\textrm{EI}}(\mathbf{x} | f^*) = \mathbb{E} \left[ \max \left(0, f^* - y \right) \right | y \sim g(\mathbf{x} | \mathcal{D}_f)].
\end{equation*}
The GP posterior $g(\mathbf{x} | \mathcal{D}_f)$ is normally distributed with known mean $\mu_f(\mathbf{x})$ and variance $\sigma^2_f(\mathbf{x})$, so this expectation has an elegant closed form in terms of the Gaussian density and distribution functions:
\begin{align}\label{eq:ei}
\alpha_{\textrm{EI}}(\mathbf{x} | f^*) = \sigma_f(\mathbf{x}) z \Phi(z) + \sigma_f(\mathbf{x}) \phi(z), \textrm{ where } z = \frac{f^* - \mu_f(\mathbf{x})}{\sigma_f(\mathbf{x})}.
\end{align}
This function is easy to implement, easy to optimize, has strong theoretical guarantees \citep{bull11}, and performs well in practice \citep{snoek12}.

\subsection{Noisy observations}\label{sec:noisy_obs}
Suppose that we do not observe $f(\mathbf{x}_i)$, rather we observe
$y_i = f(\mathbf{x}_i) + \epsilon_i$,
where $\epsilon_i$ is the observation noise, for the purposes of GP regression assumed to be $\epsilon_i \sim \mathcal{N}(0, \tau_i^2)$. Given noisy observations with uncertainty estimates $\mathcal{D}_f = \{\mathbf{x}_i, y_i, \tau_i\}_{i=1}^n$, GP regression proceeds similarly to the noiseless case and we obtain the GP posterior $g(\mathbf{x} | \mathcal{D}_f)$.

Computing EI with observation noise is challenging because we no longer know the function value of the current best point, $f^*$. \citet{gramacy10} recognize this problem and propose replacing $f^*$ in (\ref{eq:ei}) with the GP mean estimate of the best function value: $g^* = \min_{\mathbf{x}} \mu_f(\mathbf{x})$. This strategy is referred to as a ``plug-in" by \citet{picheny13}. With this substitution, EI can be computed and optimized in a similar way as in the noiseless case.

Measuring EI relative to the GP mean can be a reasonable heuristic, but when noise levels are high it can underperform. \citet{vazquez08} show that EI relative to the GP mean suffers from slow convergence to the optimum. Empirically, we found in our experiments that EI relative to the GP mean can often produce clustering of candidates and fail to sufficiently explore the space. This behavior is illustrated in Fig. S7 in the supplement.

\citet{huang06} handle this issue by defining an augmented EI which adds a heuristic multiplier to EI to increase the value of points with high predictive variance. EI is measured relative to the GP mean of the point with the best quantile, which they call the ``effective best solution." The multiplier helps to avoid over-exploitation but is not derived from any particular utility function and is primarily justified by empirical performance. \citet{picheny10, picheny13} substitute a quantile in the place of the mean for the current best, and then directly optimize expected improvement of that quantile. Quantile EI also has an analytic expression and so can be easily maximized, in their application for multi-fidelity optimization with a budget.

\citet{picheny13b} show the performance of a large collection of acquisition functions on benchmark problems with noise. The methods that generally performed the best were the augmented EI and the knowledge gradient, which is described in Section \ref{sec:other_acq}.

\subsection{Constraints}\label{sec:noisy_con}

\citet{schonlau98} extend EI to solve noiseless constrained optimization problems of the form
\begin{equation*}
\min_{\mathbf{x}} f(\mathbf{x}) \textrm{ subject to } c_j(\mathbf{x}) \leq 0, \quad j = 1, \ldots, J,
\end{equation*}
where the constraint functions $c_j(\mathbf{x})$ are also black-box functions that are observed together with $f$. As with $f$, we give each $c_j$ a GP prior and denote its posterior mean and variance as $\mu_{c_j}(\mathbf{x})$ and $\sigma^2_{c_j}(\mathbf{x})$. Let $f_c^*$ be the value of the best \textit{feasible} observation. \citet{schonlau98} define the improvement of a candidate $\mathbf{x}$ over $f_c^*$ to be $0$ if $\mathbf{x}$ is infeasible, and otherwise to be the usual improvement. Assuming independence between $f$ and each $c_j$ given $\mathbf{x}$, the expected improvement with constraints is then
\begin{equation}\label{eq:eic}
\alpha_{\textrm{EIC}}(\mathbf{x}| f_c^*) =  \alpha_{\textrm{EI}}(\mathbf{x} | f_c^*) \prod_{j=1}^J \mathbb{P}(c_j(\mathbf{x}) \leq 0).
\end{equation}
As with unconstrained EI, this quantity is easy to optimize and works well in practice \citep{gardner14}.

When the observations of the constraint functions are noisy, a similar challenge arises as with noisy observations of $f$: We may not know which observations are feasible, and so cannot compute the best feasible value $f_c^*$. \citet{gelbart14} propose using the best GP mean value that satisfies each constraint $c_j(\mathbf{x})$ with probability at least $1 - \delta_j$, for a user-specified threshold $\delta_j$ ($0.05$ in their experiments). If there is no $\mathbf{x}$ that satisfies the constraints with the required probability, then they select the candidate that maximizes the probability of feasibility, regardless of the objective value. In a high-noise setting, this heuristic for setting $f_c^*$ can be exploitative because it gives high EI for replicating points with good objective values until their probability of feasibility is driven above $1-\delta_j$.

The alternative versions of EI designed for noisy observations, described in Section \ref{sec:noisy_obs}, have not been adapted to handle constraints. Augmented EI and quantile EI, for example, require nontrivial changes to handle noisy constrants. The strategy for selecting the best observation would need to be changed to consider uncertain feasibility, and the multiplier for augmented EI would need to somehow take into account the predictive variance of the constraints.

\citet{gramacy16} describe a different approach for handling constraints in which the constraints are brought into the objective via a Lagrangian. EI is no longer analytic, but can be evaluated numerically with Monte Carlo integration over the posterior, or after reparameterization via quadrature \citep{picheny16}. The integration over the posterior naturally handles observation noise, and the same heuristics for selecting a best-feasible point can be used.

\subsection{Batch optimization}\label{sec:batch}

EI can be used for batch or asynchronous optimization by iteratively maximizing EI integrated over pending outcomes \citep{ginsbourger10}. Let $\mathbf{x}^b_1, \ldots, \mathbf{x}^b_m$ be $m$ candidates whose observations are pending, and $\mathbf{f}^b = [f(\mathbf{x}^b_1), \ldots, f(\mathbf{x}^b_m)]$ the corresponding \textit{unobserved} outcomes at those points. Candidate $m+1$ is chosen as the point that maximizes
\begin{equation}\label{eq:batch_ei}
\alpha_{\textrm{EIB}}(\mathbf{x}| f^*) = \int_{\mathbf{f}^b} \alpha_{\textrm{EI}}(\mathbf{x} | \min(f^*, \mathbf{f}^b)) p(\mathbf{f}^b | \mathcal{D}_f) d\mathbf{f}^b.
\end{equation}
Because of the GP prior on $f$, the conditional posterior $\mathbf{f}^b | \mathcal{D}_f$ has a multivariate normal distribution with known mean and covariance matrix. The integral in (\ref{eq:batch_ei}) does not have an analytic expression, but we can sample from $p(\mathbf{f}^b |  \mathcal{D}_f)$ and so can use a Monte Carlo approximation of the integral. \citet{snoek12} describe this approach to batch optimization, and show that despite the Monte Carlo integration it is efficient enough to be practically useful for optimizing machine learning hyperparameters. This approach has not previously been studied in a noisy setting.

\citet{taddy09} handle noise in batch optimization of EI by integrating over samples from the multi-point EI posterior \citep[implemented in][]{gramacy10b}. To maintain tractability, their approach is limited to evaluating EI on a discrete set of points. Here we take a similar approach and integrate over the EI posterior, but use the iterative approach in (\ref{eq:batch_ei}) to allow optimizing the integrated EI over a continuous space.

\subsection{Alternative acquisition functions}\label{sec:other_acq}
There are several other acquisition functions that handle noise more naturally than EI. One class of methods are information-based and seek to reduce uncertainty in the location of the optimizer. These methods include IAGO \citep{villemonteix09}, entropy search \citep{hennig12}, and predictive entropy search (PES) \citep{hernandez14}. Predictive entropy search has been developed to handle constraints \citep{hernandez15} and batch optimization \citep{shah15}. Although the principle behind PES is straightforward (select the point that most reduces predictive entropy of the location of the minimizer), the quantities that must be calculated are intractable and a collection of difficult-to-implement approximations must be used.

Another acquisition function that naturally handles noise is the knowledge gradient \citep{scott11}.  Knowledge gradient has been extended to batch optimization \citep{wu16, wang16}, but has not been extended to constrained problems.  Optimizing the knowledge gradient repeatedly is the myopic one-step optimal policy, and each optimization selects the point that will be most useful in expectation if the next decision is to select the best point.  Constraints cannot be simply added to the knowledge gradient without losing the tractability of this expectation, and the construction of a knowledge gradient suitable for noisy constraints would involve a substantial update to the implicit procedure for selecting the best point.

Recently the classic Thompson sampling algorithm \citep{thompson33} has been applied to GP Bayesian optimization \citep{hernandez17, kandasamy18}. This approach optimizes the objective on individual draws from the GP posterior to provide highly parallel optimization.

\subsection{Selecting the best point after Bayesian optimization}
The final step of Bayesian optimization, referred to as the identification step by \citet{jalali17}, is to decide which evaluated point is best.  Without noise this step is trivial, but with noise a difficult decision must be made.  For noisy objectives without constraints, typical strategies are to choose the point with the best GP mean or the best quantile \citep{jalali17}.

For A/B tests where the choice of best point can have longstanding effects, teams often prefer to manually select the best point according to their understanding of the trade-offs between constraints, objectives, and uncertainty.

For closed-loop optimization or other settings where a rigid criterion is required, one approach is to select the point that has the largest expected reduction in objective over a known baseline $B$, which could be the objective achieved by a worst-case (i.e. largest) feasible objective value. This is the point maximizing 
\begin{equation}\label{eq:identification}
\left( B - \mu_f(\mathbf{x}) \right) \prod_{j=1}^J \mathbb{P}(c_j(\mathbf{x}) \leq 0)
\end{equation}
over the evaluated points. Another approach is to select the point that has the smallest posterior mean objective that meets all constraints, or each constraint, with probability $1-\delta$ for a given $\delta$ \citep{gelbart14}. In our experiments we show results for both of these strategies.

\section{Utility maximization and EI with noise}\label{sec:unified}
EI is the strategy that myopically maximizes a particular utility function. By considering that utility function in the case of noisy observations and constraints we can derive an appropriate form of EI without heuristics, and will see that it extends immediately to handle asynchronous optimization. The result will be an integral similar to that of (\ref{eq:batch_ei}), but in Section \ref{sec:qmc} we develop a more efficient estimate than has previously been used for batch optimization.

\subsection{Infeasibility in the noiseless setting}
We build up from the noiseless case, where both objective and constraints are observed exactly. We begin by defining a utility function that gives the utility after $n$ iterations of optimization. To correctly deal with noisy constraints later, we must explicitly consider the case where no observations are feasible. Let $S = \{i : c_j(\mathbf{x}_i) \leq 0 \; \forall j\}$ be the set of feasible observations. The utility function is
\begin{equation*}
u(n) = 
\begin{cases}
-\min_{i \in S} f(\mathbf{x}_i) & \textrm{if } |S| > 0,\\
-M & \textrm{otherwise}.
\end{cases}
\end{equation*}
Here $M$ is a penalty for not having a feasible solution.\footnote{This penalty should be high enough that we prefer finding a feasible solution to not having a feasible solution. This can be achieved by setting $M$ greater than the largest GP estimate for the objective in the design space. The value is only important in settings where there are no feasible observations; see the supplement for further discussion on sensitivity.}
As before, $f_c^*$ is the objective value of the best feasible point after $n$ iterations. We only gain utility from points that we have observed, inasmuch as we would typically not consider launching an unobserved configuration. Note that this is the utility implied by the constrained EI formulations of \citet{schonlau98} and \citet{gardner14}. The improvement in utility from iteration $n$ to iteration $n+1$ is
\begin{align*}
I(\mathbf{x}_{n+1}) &= u(n+1) - u(n)\\
&=
\begin{cases}
0 & \mathbf{x}_{n+1} \textrm{ infeasible},\\
M - f(\mathbf{x}_{n+1}) & \mathbf{x}_{n+1} \textrm{ feasible}, S_n = \varnothing,\\
\max(0, f_c^* - f(\mathbf{x}_{n+1})) & \mathbf{x}_{n+1} \textrm{ feasible}, |S_n| > 0.
\end{cases}
\end{align*}
We choose $\mathbf{x}_{n+1}$ to maximize the expected improvement under the posterior distributions of $f(\mathbf{x})$ and $c_j(\mathbf{x})$. For convenience, let $\mathbf{f}^n = [f(\mathbf{x}_1), \ldots, f(\mathbf{x}_n)]$ be the objective values at the observations, $\mathbf{c}_j^n = [c_j(\mathbf{x}_1), \ldots, c_j(\mathbf{x}_n)]$ the values for each constraint, and $\mathbf{c}^n = [\mathbf{c}_1^n, \ldots, \mathbf{c}_J^n]$ all constraint observations. In the noiseless setting, $\mathbf{f}^n$ and $\mathbf{c}^n$ are known, the best feasible value $f_c^*$ can be computed, and the \textit{EI with infeasibility} is
\begin{align}\nonumber
\alpha_{\textrm{EIx}}(\mathbf{x} |  \mathbf{f}^n, \mathbf{c}^n) &= \mathbb{E}_{f(\mathbf{x}), c_1(\mathbf{x}), \ldots, c_J(\mathbf{x})}[I(\mathbf{x} ) | \mathbf{f}^n, \mathbf{c}^n] \\\label{eq:eix}
& = 
\begin{cases}
\alpha_{\textrm{EI}}(\mathbf{x} | f_c^*) \prod_{j=1}^J \Phi \left( -\frac{\mu_{c_j}(\mathbf{x})}{\sigma_{c_j}(\mathbf{x})} \right) & |S_n| > 0,\\
(M - \mu(\mathbf{x})) \prod_{j=1}^J \Phi \left( -\frac{\mu_{c_j}(\mathbf{x})}{\sigma_{c_j}(\mathbf{x})} \right) & \textrm{otherwise}.
\end{cases}
\end{align}
This extends the constrained EI of (\ref{eq:eic}) to explicitly handle the case where there are no feasible observations. Without a feasible best, this acquisition function balances the expected objective value with the probability of feasibility, according to the penalty $M$. As $M$ gets large, it approaches the strategy of \citet{gelbart14} and maximizes the probability of feasibility. For finite $M$, however, given two points with the same probability of being feasible, this acquisition function will choose the one with the better objective value.

\subsection{Noisy EI}

We now extend the expectation in (\ref{eq:eix}) to noisy observations and noisy constraints. This is done exactly by iterating the expectation over the posterior distributions of $\mathbf{f}^n$ and $\mathbf{c}^n$ given their noisy observations. Let $\mathcal{D}_{c_j}$ be the noisy observations of the constraint functions, potentially with heteroscedastic noise. Then, by their GP priors and assumed independence,
\begin{align*}
\mathbf{f}^n | \mathcal{D}_f &\sim \mathcal{N}(\boldsymbol{\mu}_f, \Sigma_f)\\
\mathbf{c}_j^n | \mathcal{D}_{c_j} &\sim \mathcal{N}(\boldsymbol{\mu}_{c_j}, \Sigma_{c_j}), \; j=1, \ldots, J.
\end{align*}
These are the GP posteriors for the true (noiseless) values of the objective and constraints at the observed points. The means and covariance matrices of these posterior distributions have closed forms in terms of the GP kernel function and the observed data \citep{rasmussen06}. Let $\mathcal{D} = \{\mathcal{D}_f, \mathcal{D}_{c_1}, \ldots, \mathcal{D}_{c_J}\}$ denote the full set of data. \textit{Noisy expected improvement} (NEI) is then:
\begin{align}\label{eq:full_ei}
\alpha_{\textrm{NEI}}(\mathbf{x} | \mathcal{D}) = \int_{\mathbf{f}^n} \int_{\mathbf{c}^n} \alpha_{\textrm{EIx}}& (\mathbf{x} |   \mathbf{f}^n, \mathbf{c}^n) p(\mathbf{f}^n | \mathcal{D}_f) \prod_{j=1}^J p(\mathbf{c}_j^n | \mathcal{D}_{c_j}) d\mathbf{c}^n d\mathbf{f}^n.
\end{align}
This acquisition function does not have an analytic expression, but we will show in the next section that both it and its gradient can be efficiently estimated, and so it can be optimized.

This approach extends directly to allow for batch or asynchronous optimization with noise and constraints following the approach of Section \ref{sec:batch}. The objective values at the observed points, $\mathbf{f}^n$, and at the earlier points in the batch, $\mathbf{f}^b$, are jointly normally distributed with known mean and covariance. The integral in (\ref{eq:full_ei}) is over the true values of all previously sampled points. For batch optimization, we simply extend that integral to be over both the previously sampled points and over any pending observations. Replacing $\mathbf{f}^n$ in (\ref{eq:full_ei}) with $[\mathbf{f}^n, \mathbf{f}^b]$ and making the corresponding replacement for $\mathbf{c}^n$ yields the formula for batch optimization.

Without observation noise, NEI is exactly EI. Like EI in the noiseless setting, NEI is always 0 at points that have already been observed and so will never replicate points. Replication can generally be valuable for reducing uncertainty at a possibly-good point, although with the GP we can reduce uncertainty at a point by sampling points in its neighborhood. NEI will typically sample many points near the optimum to reduce uncertainty at the optimum without having to replicate. This behavior is illustrated in Fig. S7 in the supplement, which shows the NEI candidates from an optimization run of Section \ref{sec:ei_sim}.

\section{Efficient quasi-Monte Carlo integration of noisy EI}\label{sec:qmc}

For batch optimization in the noiseless unconstrained case, the integral in (\ref{eq:batch_ei}) is estimated with Monte Carlo (MC) sampling. The dimensionality of that integral equals the number of pending observations. The dimensionality of the NEI integral in (\ref{eq:full_ei}) is the total number of observations, both pending and completed. We benefit from a more efficient integral approximation, for which we turn to quasi-Monte Carlo (QMC) methods.

QMC methods provide an efficient approximation of high-dimensional integrals on the unit cube as a sum of function evaluations:
\begin{equation*}
\int_{[0, 1]^d} f(\mathbf{u}) d\mathbf{u} \approx \frac{1}{N} \sum_{k=1}^N f(\mathbf{t}_k).
\end{equation*}
When $\mathbf{t}_k$ are chosen from a uniform distribution on $[0, 1]^d$, this is MC integration. The Central Limit Theorem provides a convergence rate of $\mathcal{O}(1/\sqrt{N})$ \citep{caflisch98}. QMC methods provide faster convergence and lower error by using a better choice of $\mathbf{t}_k$. For the purposes of integration, random samples can be wasteful because they tend to clump; a point that is very close to another provides little additional information about a smooth $f$. QMC methods replace random samples for $\mathbf{t}_k$ with a deterministic sequence that is constructed to be low-discrepancy, or space-filling. There are a variety of such sequences, and here we use Sobol sequences \citep{owen98}. Theoretically, QMC methods achieve a convergence rate of $\mathcal{O}((\log N)^ d / N)$, and typically achieve much faster convergence in practice \citep{dick13}. The main theoretical result for QMC integration is the Koksma-Hlawka theorem, which provides a deterministic bound on the integration error in terms of the smoothness of $f$ and the discrepancy of $\mathbf{t}_k$ \citep{caflisch98}.

To use QMC integration to estimate the NEI in (\ref{eq:full_ei}), we must transform that integral to the unit cube.
\begin{proposition}[\citealp{dick13}]\label{prop:1}
Let $p(\mathbf{x} | \boldsymbol{\mu}, \Sigma)$ be the multivariate normal density function and choose $A$ such that $\Sigma = AA^\intercal$. Then,
\begin{equation*}
\int_{\mathbb{R}^d} f(\mathbf{y}) p(\mathbf{y} | \boldsymbol{\mu}, \Sigma)  d\mathbf{y} = \int_{[0, 1]^d} f(A\Phi^{-1}(\mathbf{u}) + \boldsymbol{\mu}) d\mathbf{u}.
\end{equation*}
\end{proposition}
The matrix $A$ can be the Cholesky decomposition of $\Sigma$. We now apply this result to the NEI integral in (\ref{eq:full_ei}).
\begin{proposition}
Let $\Sigma = \textrm{diag}(\Sigma_f, \Sigma_{c_1}, \ldots, \Sigma_{c_J})$ and $\boldsymbol{\mu} = [\boldsymbol{\mu}_f, \boldsymbol{\mu}_{c_1}, \ldots, \boldsymbol{\mu}_{c_J}]$. Choose $A$ such that $\Sigma = AA^\intercal$ and let
\begin{equation*}
\begin{bmatrix}
\tilde{\mathbf{f}}^n(\mathbf{u})\\
\tilde{\mathbf{c}}^n(\mathbf{u})\\
\end{bmatrix}
= A\Phi^{-1}(\mathbf{u}) + \boldsymbol{\mu},
\end{equation*}
with $\tilde{\mathbf{f}}^n(\mathbf{u}) \in \mathbb{R}^n$ and $\tilde{\mathbf{c}}^n(\mathbf{u}) \in \mathbb{R}^{Jn}$. Then,
\begin{equation*}
\alpha_{\textrm{NEI}}(\mathbf{x} | \mathcal{D}) = \int_{[0, 1]^{n(J+1)}} \alpha_{\textrm{EIx}}(\mathbf{x} | \tilde{\mathbf{f}}^n(\mathbf{u}), \tilde{\mathbf{c}}^n(\mathbf{u})) d\mathbf{u}.
\end{equation*}
\end{proposition}
QMC methods thus provide an estimate for the NEI integral according to
\begin{equation}\label{eq:qmc_ei}
\alpha_{\textrm{NEI}}(\mathbf{x} | \mathcal{D}) \approx \frac{1}{N} \sum_{k=1}^N \alpha_{\textrm{EIx}}(\mathbf{x} | \tilde{\mathbf{f}}^n(\mathbf{t}_k), \tilde{\mathbf{c}}^n(\mathbf{t}_k)).
\end{equation}
The transform $A\Phi^{-1}(\mathbf{u}) + \boldsymbol{\mu}$ is the typical way that multivariate normal random samples are generated from uniform random samples $\mathbf{u}$ \citep{gentle09}. Thus when each $\mathbf{t}_k$ is chosen uniformly at random, this corresponds exactly to Monte Carlo integration using draws from the GP posterior. Using a quasirandom sequence $\{\mathbf{t}_1, \ldots, \mathbf{t}_N\}$ provides faster convergence, and so reduces the number of samples $N$ required for optimization.

\begin{figure}[t]
\centering
\includegraphics{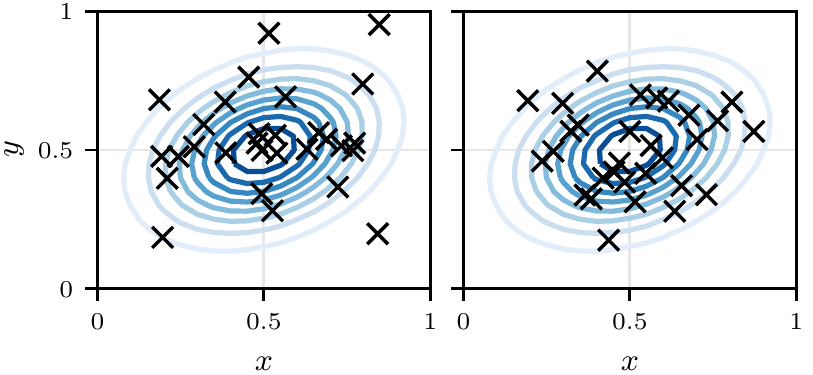}
\caption{(Left) Multivariate normal random samples. (Right) Space-filling quasirandom multivariate normal samples.}
\label{fig:qmc_example}
\end{figure}

As an illustration, Fig. \ref{fig:qmc_example} shows random draws from a multivariate normal alongside quasirandom ``draws" from the same distribution, generated by applying the transform of Proposition \ref{prop:1} to a scrambled Sobol sequence. The quasirandom samples have better coverage of the distribution and will provide lower integration error.

The algorithm for computing NEI is summarized in Algorithm \ref{algo:nei}. In essence, we draw QMC samples from the posteriors for the true values of the noisy observations, and for each sampled ``true" value, we compute noiseless EI using (\ref{eq:eix}). The computationally intensive steps in Algorithm \ref{algo:nei} are kernel inference in line 1 and constructing the noiseless GP models in line 8. For the noiseless GP models we reuse the kernel hyperparameters from line 1, but must still invert each of their covariance matrices. Lines 1--8 (the QMC sampling and constructing the noiseless models for each sample) are independent of the candidate $\mathbf{x}$. In practice, we do these steps once at the beginning of the optimization and cache the models.  When we wish to evaluate the expected improvement at any point $\mathbf{x}$ during the optimization, we evaluate the GP posteriors at $\mathbf{x}$ for each of these cached models and compute EI (lines 10--13). This allows NEI to be quickly computed and optimized. For asynchronous or batch optimization, the posteriors in line 2 are those of both completed and pending observations, and all other steps remain the same.  Note that line 3 utilizes the assumed independence of the objective and constraint values from line 2, but the algorithm could utilize a full covariance matrix across functions if available.

\begin{algorithm}[htb]\label{algo:nei}
 \KwData{Noisy objective and constraint observations $\mathcal{D}$, candidate $\mathbf{x}$.}
 \KwResult{Expected improvement at $\mathbf{x}$.}
 Infer GP kernel hyperparameters for objective and constraints, from $\mathcal{D}$.
 
 Compute GP posteriors for the objective and constraint values at the observations:
\begin{align*}
\mathbf{f}^n | \mathcal{D}_f &\sim \mathcal{N}(\boldsymbol{\mu}_f, \Sigma_f)\\
\mathbf{c}_j^n | \mathcal{D}_{c_j} &\sim \mathcal{N}(\boldsymbol{\mu}_{c_j}, \Sigma_{c_j}), \; j=1, \ldots, J.
\end{align*}

Construct $\Sigma = \textrm{diag}(\Sigma_f, \Sigma_{c_1}, \ldots, \Sigma_{c_J})$ and $\boldsymbol{\mu} = [\boldsymbol{\mu}_f, \boldsymbol{\mu}_{c_1}, \ldots, \boldsymbol{\mu}_{c_J}]$.

Compute the Cholesky decomposition $\Sigma = AA^\intercal$.

Generate a quasi-random sequence $\mathbf{t}_1, \ldots, \mathbf{t}_N$.

\For{$i = 1, \ldots, N$}{

Draw quasi-random samples from the GP posterior for the values at the observations:
\begin{equation*}
\begin{bmatrix}
\tilde{\mathbf{f}}_i\\
\tilde{\mathbf{c}}_i\\
\end{bmatrix}
= A\Phi^{-1}(\mathbf{t}_i) + \boldsymbol{\mu}.
\end{equation*}

Construct a GP model $\mathcal{M}_i$ with noiseless observations $\tilde{\mathbf{f}}_i$ and $\tilde{\mathbf{c}}_i$.
}

Initialize $\alpha_{\textrm{NEI}} = 0$.

\For{$i = 1, \ldots, N$}{
Compute the posterior at $\mathbf{x}$ under model $\mathcal{M}_i$.

Use this GP posterior to compute EI as in the noiseless setting, $\alpha_{\textrm{EIx}}$ in (\ref{eq:eix}).

Increment $\alpha_{\textrm{NEI}} = \alpha_{\textrm{NEI}} + \frac{1}{N}\alpha_{\textrm{EIx}}$.
}

\Return{$\alpha_{\textrm{NEI}}$}
\caption{Noisy EI with QMC integration}
\end{algorithm}

The gradient of $\alpha_{\textrm{EIx}}$ can be computed analytically, and so the gradient of (\ref{eq:qmc_ei}) is available analytically and NEI can be optimized with standard nonlinear optimization methods. Besides the increased dimensionality of the integral, it is no harder to optimize (\ref{eq:qmc_ei}) than it is to optimize (\ref{eq:batch_ei}), which has been shown to be efficient enough for practical use. Optimizing (\ref{eq:batch_ei}) for batch EI requires sampling from the GP posterior and fitting conditional models for each sample just as in Algorithm \ref{algo:nei}. We now show that the QMC integration allows us to handle the increased dimensionality of the integral and makes NEI practically useful.

\section{Synthetic problems}\label{sec:simulations}
We use synthetic problems to provide a rigorous study of two aspects of our method. In Section \ref{sec:qmc_sim} we compare the performance of QMC integration to the MC approach used to estimate (\ref{eq:batch_ei}). We show that QMC integration allows the use of many fewer samples to achieve the same integration error and optimization performance, thus allowing us to efficiently optimize NEI. In Section \ref{sec:ei_sim} we compare the optimization performance of NEI to that of several baseline approaches, and show that NEI significantly outperformed the other methods.

We used four synthetic problems for our study. The equations and visualizations for each problem are given in the supplement. The first problem comes from \citet{gramacy16}, and has two parameters and two constraints. The second is a constrained version of the Hartmann 6 problem with six parameters and one constraint, as in \citet{jalali17}. The third problem is a constrained Branin problem used by \citet{gelbart14} and the fourth is a problem given by \citet{gardner14}; these both have two parameters and one constraint. We simulated noisy objective and constraint observations by adding normally distributed noise to evaluations of the objective and constraints. Noise variances for each problem are given in the supplement.

In the experiments here and in Section \ref{sec:real-world}, GP regression was done using a Mat{\'e}rn 5/2 kernel, and posterior distributions for the kernel hyperparameters were inferred using the NUTS sampler \citep{hoffman14}. GP predictions were made using the posterior mean value for the hyperparameters. NEI was optimized using random restarts of the Scipy SLSQP optimizer. In a typical randomized experiment, including those of Section \ref{sec:real-world}, we observe both the mean estimate and its standard error. All methods were thus given the true noise variance of each observation.

\subsection{Evaluating QMC performance}\label{sec:qmc_sim}

The first set of simulations analyze the performance of the QMC estimate in (\ref{eq:qmc_ei}). We simulated computing NEI in a noisy, asynchronous setting by using observations at 5 quasirandom points as data, and then treating an additional 5 quasirandom points as pending observations. We then estimated the NEI integral of (\ref{eq:full_ei}) at a point using regular MC draws from the posterior, and using QMC draws as in Algorithm \ref{algo:nei}. The locations of these points and the true NEI surfaces are given in Fig. S5 in the supplement.

\begin{figure}[tb]
\centering
\includegraphics{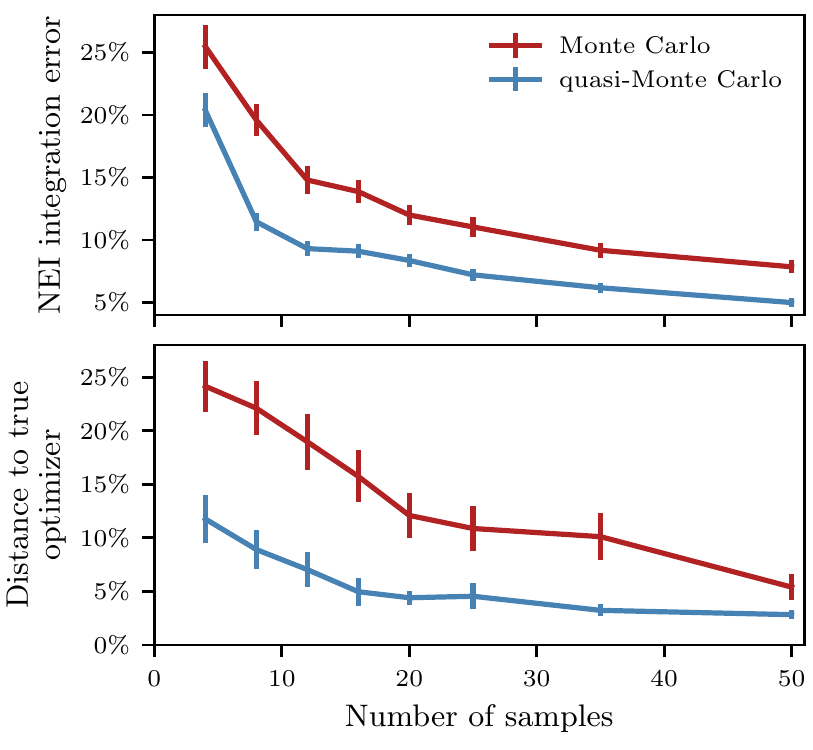}
\caption{(Top) NEI integration error (average over 500 replicates, and two standard errors of the mean) as a function of the number of MC or QMC samples used for the approximation. (Bottom) Average distance from the optimizer using the approximated NEI to the true NEI global optimum, as a percent of the maximum distance in the search space. QMC yielded substantially better optimization performance.}
\label{fig:qmc_sim}
\end{figure}

For a range of the number of MC and QMC samples, we measured the percent error relative to the ground-truth found by estimating NEI with $10^4$ regular MC samples. Fig. \ref{fig:qmc_sim} shows the results for the Gramacy problem. For this problem, QMC reliably required half as many samples as MC to achieve the same integration error.

Typically we are not interested in the actual value of NEI, rather we only want to find the optimizer. For 100 replicates, we optimized NEI using the MC and QMC approximations, and measured the Euclidean distance between the found optimizer and the ground-truth optimizer. Fig. \ref{fig:qmc_sim} shows that the lower integration error led to better optimization performance: 16 QMC samples achieved the same optimizer distance as 50 MC samples. This same simulation was done for the other three problems, and similar results are shown in Fig. S6 in the supplement.

\subsection{Optimization performance compared to heuristics and other methods}\label{sec:ei_sim}

We compare optimization performance of NEI to using the heuristics of Section \ref{sec:prior} to handle the noise in observations and constraints and to available baselines. For the EI+heuristics method, we measure expected improvement relative to the best GP mean of points that satisfy the constraints in expectation. Batch optimization is done as described in Section \ref{sec:batch}, but using MC draws from a GP that includes the observation noise. The EI+heuristics method uses the same GP models and optimization routines as the NEI method, with the only difference being the use of heuristics in computing EI. In particular, the methods are identical in the absence of observation noise. In addition to the heuristics baseline, we also compare to two commonly used Bayesian optimization methods from the Spearmint package: Spearmint EI \citep{snoek12}, and Spearmint PESC \citep{hernandez15}. Spearmint EI uses similar heuristics as EI+heuristics to handle noise, but also uses a different approach for GP estimation, different optimization routines, and other techniques like input warping \citep{snoek14}. Spearmint PESC implements constrained predictive entropy search. There are a number of other available packages for Bayesian optimization, however only Spearmint currently supports constraints and so our comparison is limited to these methods.

Each optimization was begun from the same batch of 5 Sobol sequence points, after which Bayesian optimization was performed in 9 batches of 5 points each, for a total of 50 iterations. After each batch, noisy observations of the points in the batch were incorporated into the model. This simulation was repeated 100 times for each of the four problems, each with independent observation noise added to function and constraint evaluations.

\begin{figure*}[tb]
\includegraphics{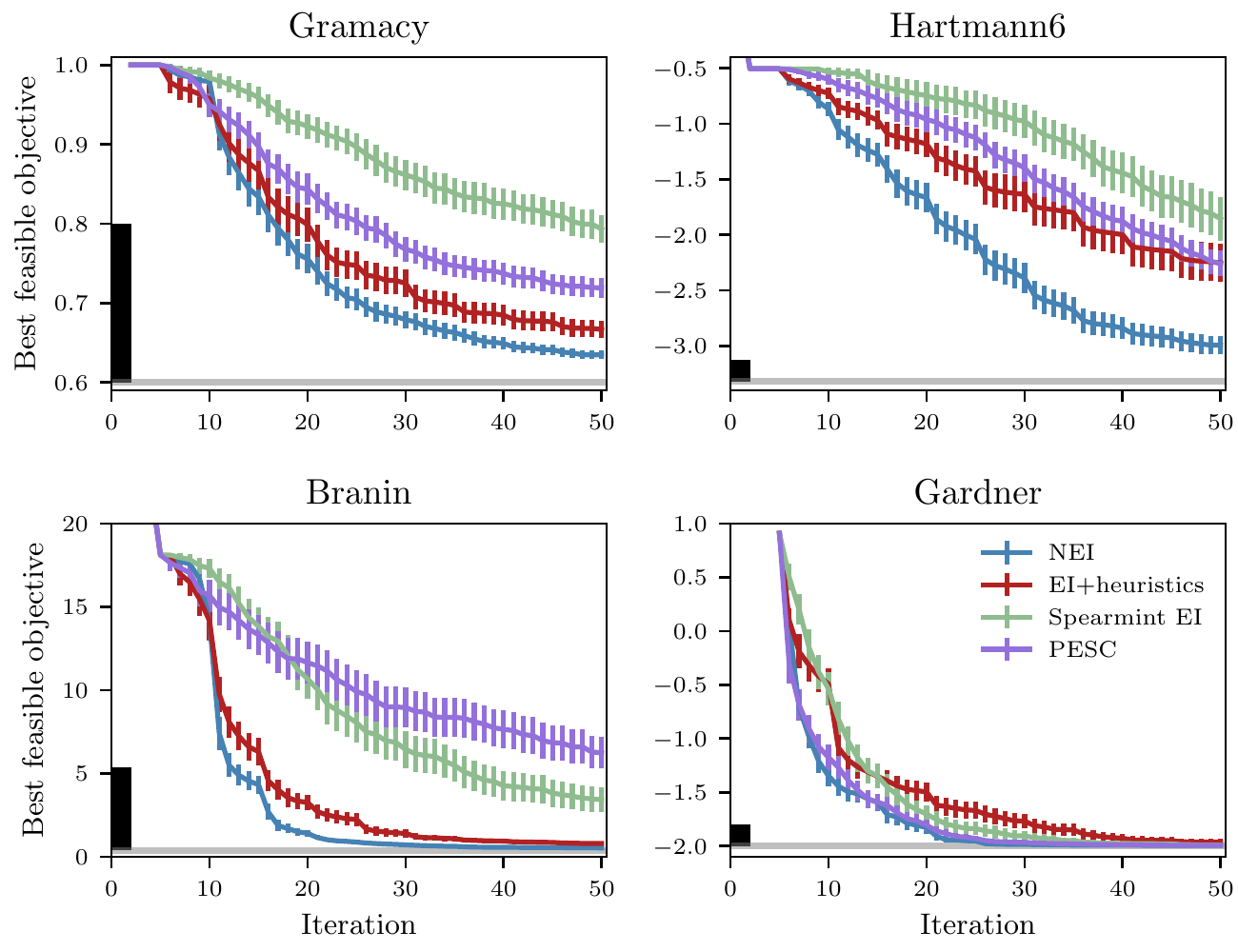}
\caption{Value of the best feasible objective by each iteration of optimization, for each of the four problems and each of the four methods. Plots show mean over replicates and two standard errors of the mean. Horizontal line indicates the global optimum for the problem and the black bar is the standard deviation of the observation noise. NEI consistently outperformed the other methods.}
\label{fig:bo_sim}
\end{figure*}

Fig. \ref{fig:bo_sim} shows the value of the best feasible point at each iteration of the optimization, for all four problems. NEI consistently performed the best of all of the methods. Compared to EI+heuristics, NEI was able to find better solutions with fewer iterations. Without noise, these two methods are identical; the improved performance comes entirely from correctly handling observation noise. PESC had equal performance as NEI on the Gardner problem, but performed worse even than EI+heuristics on the other problems. Computation time was similar for the four methods, all requiring around 10s per iteration.

\begin{figure}[tb]
\centering
\includegraphics{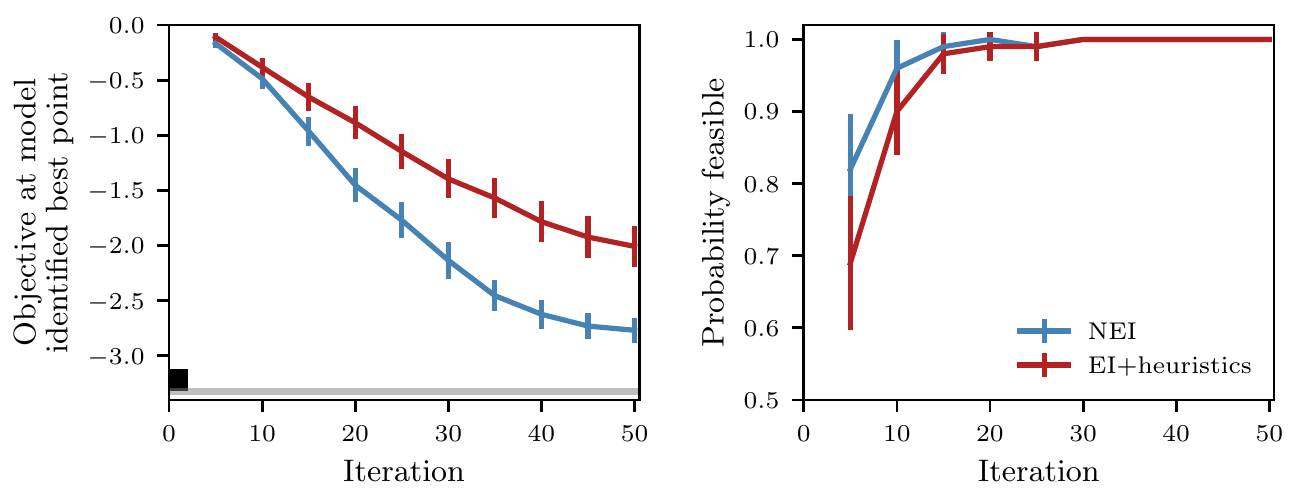}
\caption{(Left) For the Hartmann6 problem, the objective value of the arm identified from the model as being best after each batch of the simulation in Fig. \ref{fig:bo_sim}. (Right) The proportion of replicates in which the model identified best point was actually feasible. NEI was able to both find and identify better points.}
\label{fig:model_best}
\end{figure}

As illustrated in Fig. S7 in the supplement, the proposals from EI+heuristics tended to form clumps at points with a good objective value and uncertain feasibility. Being more exploitative in a noisy setting could potentially be advantageous by allowing the model to more accurately identify the best feasible solution. We compare the final model identified best points after each batch for NEI and EI+heuristics for the Hartmann6 problem in Fig. \ref{fig:model_best}, according to the criterion of (\ref{eq:identification}). By the final batch of the optimization, both methods were able to identify arms that were feasible but those chosen by NEI had significantly better objective. Similar results for the other three problems are given in Fig. S9 of the supplement. Fig. S10 of the supplement shows results using the alternative identification strategy of choosing the best arm that is feasible with probability greater than $1-\delta$.

\section{Bayesian optimization with real-world randomized experiments}\label{sec:real-world}
We present two case studies of how Bayesian optimization with NEI works in practice with real experiments at Facebook: an online field experiment to optimize ranking system parameters, and a randomized controlled benchmark to optimize server performance. Both experiments involved tuning many continuous parameters simultaneously via noisy objectives and noisy constraints.

\subsection{Optimizing machine learning systems}
Advances in modeling, feature engineering, and hyperparameter optimization are typical targets for improving the performance of the models that make up a machine learning system. However, the performance of a machine learning system also depends on the inputs to the model, which often come from many interconnected retrieval and ranking systems, each of which is controlled by many tuning parameters \citep{bendersky2010anatomy,covington2016deep}. For example, an indexer may retrieve a subset of items which are then fed into a high-precision ranking algorithm. The indexer has parameters such as the number of items to retrieve at each stage and how different items are valued \citep{rodriguez2012multiple}. Tuning these parameters can often be as important as tuning the model itself.

While Bayesian optimization has proven to be an effective tool for optimizing the performance of machine learning models operating in isolation \citep{snoek12}, the evaluation of an entire production system requires live A/B testing.  Since outcomes directly affected by machine learning systems are heavily skewed \citep{kohavi2014seven}, measurement error is on the same order as the effect size itself.

We used NEI to optimize a ranking system. This system consisted of an indexer that aggregated content from various sources and identified items to be sent to a model for ranking. We experimented with tuning indexer parameters in a 6-dimensional space to improve the overall performance of the system. We maximized an objective metric subject to a lower bound on a constraint metric. NEI is ideally suited for this type of randomized experiment: noise levels are significant relative to the effect size, multiple variants are tested simultaneously in a batch fashion, and there are constraints that must be satisfied (e.g., measures of quality).

\begin{figure}[tbh]
\centering
\includegraphics{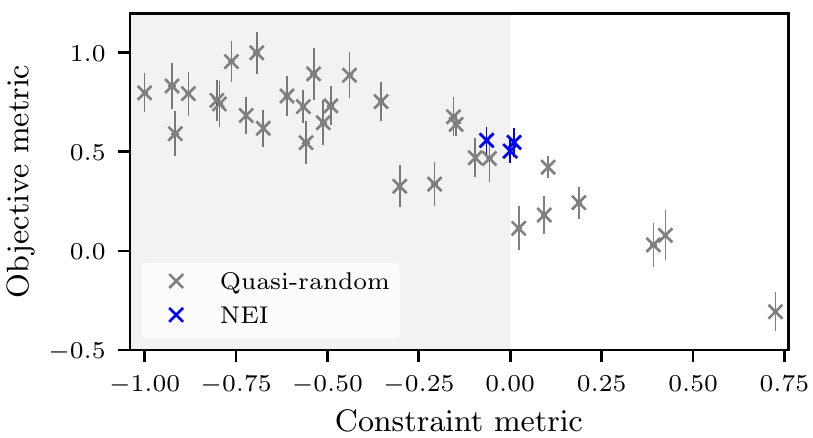}
\caption{Posterior GP predictions (means and 2 standard deviations) from an A/B test using NEI to generate a batch of 3 candidates. The goal was to maximize the objective, subject to a lower bound on the constraint. The shaded region is infeasible. NEI found a feasible point with significantly better objective value than both the baseline and the quasirandom initialization.}
\label{fig:live_experiment}
\end{figure}

The experiment was conducted in two batches: a quasirandom initial batch of 31 configurations selected with a scrambled Sobol sequence, and a second batch which used NEI to propose 3 configurations. Fig.~\ref{fig:live_experiment} shows the results of the experiment as change relative to baseline, with axes scaled by the largest effect. In this experiment, the objective and constraint were highly negatively correlated ($\rho = 0.78$). NEI proposed candidates near the constraint boundary, and with only three points was able to find a feasible configuration that improved over both the baseline and anything from the initial batch. 

\subsection{Optimizing server performance}

We applied Bayesian optimization with NEI to improve the performance of the servers that power Facebook. Facebook is written in a mix of the PHP and Hack programming languages, and it uses the HipHop Virtual Machine~(HHVM)~\citep{adams14} to execute the PHP/Hack code in order to serve HTTP requests.  HHVM is an open-source virtual machine containing a just-in-time~(JIT) compiler to translate the PHP/Hack code into Intel x86 machine code at runtime so it can be executed.  

During the compilation process, HHVM's JIT compiler performs a large number of code optimizations aimed at improving the performance of the final machine code. For example, code layout optimization splits the hot and cold code paths in order to improve the effectiveness of the instruction cache by increasing the chances of the hot code remaining in the cache.  How often a code block is
executed to be considered hot is a tunable parameter inside the JIT compiler. As another example, function inlining eliminates the overhead of calling and returning from a function, with tunable parameters determining which kinds of functions should be inlined.

Tuning compiler parameters can be very challenging for a number of reasons. First, even seemingly unrelated compiler optimizations, such as function inlining and code layout, can interfere with one another by affecting performance of the processor's instruction cache. Second, there are often additional constraints that limit the viable optimization space.  Function inlining, for example, can drastically increase code size and, as a result, memory usage. Third, accurate modeling of all the factors inside a processor is so difficult that the only reasonable way to compare the performance of two different configurations is by running A/B tests. 

Facebook uses a system called Perflab for running A/B tests of server configurations \citep{Bakshy15}. At a high-level, a Perflab experiment assigns two isolated sets of machines to utilize the two configurations. It then replays a representative sample of user traffic against these hosts at high load, while measuring performance metrics including CPU time, memory usage, and database fetches, among other things. Perflab provides confidence intervals on these noisy measurements, characterizing the noise level and allowing for rigorous comparison of the configurations. The system is described in detail in \citet{Bakshy15}. Each A/B traffic experiment takes several hours to complete, however we had access to several machines on which to run these experiments, and so could use asynchronous optimization to run typically 3 function evaluations in parallel.

We tuned $7$ numeric run-time compiler flags in HHVM that control inlining and code layout optimizations. This was a real experiment that we conducted, and the results were incorporated into the mainstream open-source HHVM~\citep{ottoni:16:github}. Parameter names and their ranges are given in the supplement. Some parameters were integers---these values were rounded after optimization for each proposal. The goal of the optimization was to reduce CPU time with a constraint of not increasing peak memory usage on the server.

\begin{figure*}[tb]  
\centering
\includegraphics{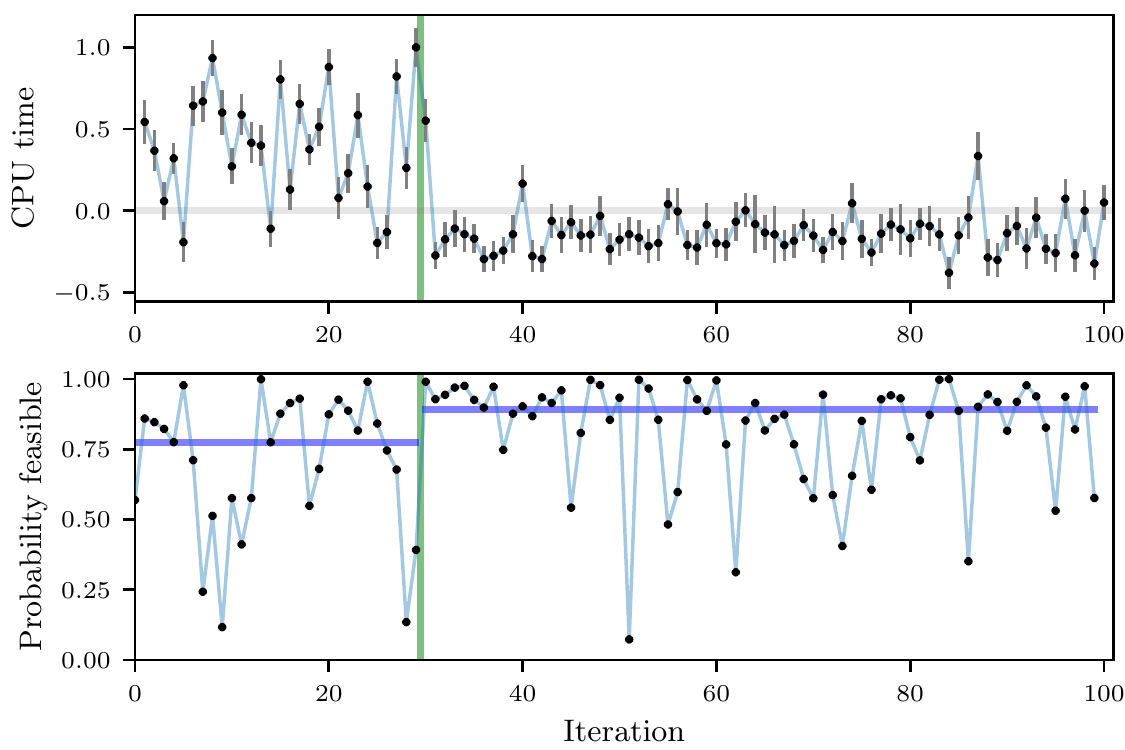}
\caption{(Left) Posterior GP predictions (means and 2 standard deviations) of CPU time across the optimization iterations, as scaled change relative to baseline. The vertical line marks the end of the quasirandom initialization and the start of candidates selected using NEI. The objective was to minimize CPU time, subject to peak memory not increasing. (Right) The probability of feasibility at each iteration. Horizontal lines show the median for the quasirandom points and for the NEI points. NEI candidates reduced CPU time and increased probability of feasibility.}
\label{fig:hhvm_iterations}
\end{figure*}

We initialized with $30$ configurations that were generated via scrambled Sobol sequences and then ran $70$ more traffic experiments whose configurations were selected using NEI. Fig.~\ref{fig:hhvm_iterations} shows the CPU time and probability of feasibility across iterations. In the quasirandom initialization, CPU time and memory usage were only weakly correlated ($\rho = 0.21$). CPU times shown were scaled by the maximum observed difference. The optimization succeeded in finding a better parameter configuration, with experiment $83$ providing the most reduction in CPU time while also not increasing peak memory. Nearly all of the NEI candidates provided a reduction of CPU time relative to baseline, while also being more likely to be feasible: the median probability of feasibility in the initialization was 0.77, which increased to 0.89 for the NEI candidates.

\section{Discussion}

Properly handling noisy observations and noisy constraints is important when tuning parameters of a system via sequential experiments with measurement error. If the measurement error is small relative to the effect size, Bayesian optimization using a heuristic EI can be successful. However, when the measurement noise is high we can substantially improve performance by properly integrating out the uncertainty.

NEI requires solving a higher dimensional integral than has previously been used for batch optimization, but we developed a QMC integration technique which allowed the integral to be estimated efficiently enough for optimization. Even in the noiseless case, the QMC approach that we developed here could be used to speed up the batch optimization strategy of \citet{snoek12}. QMC provided a useful approximation to the integral with a relatively low number of samples. Part of the success of QMC for the NEI integral likely comes from the low effective dimensionality of this integral \citep{wang03}. The EI at a point is largely determined by the values at nearby points and at the best point. Points that are far away and not likely to be the best will have little influence on the NEI integral, and so the effective dimensionality is lower than the total number of observations.

Qualitatively, we are measuring EI under various possible realizations of the true function. Averaging over a number of such realizations finds points that have high EI under many possible true functions, which is a desirable property even if there are too few QMC samples to fully characterize the posterior. Regardless of the number of QMC samples or dimensionality of the integral, points with positive NEI estimated via sampling are guaranteed to actually have positive NEI, hence we can expect the optimization to progress.

Measuring EI at $x$ relative to the GP mean at the best $x^*$, as EI+heuristics does, ignores the covariance between $f(x)$ and $f(x^*)$. Given two points $x_1$ and $x_2$ with the same marginal posteriors $f(x_1) = f(x_2)$, we should prefer the point that is less correlated with $f(x^*)$ since our expected total utility will be higher. NEI incorporates covariance between points and so would prefer the less correlated point, whereas for EI+heuristics they would be considered equally valuable.

The NEI acquisition function does not give value to replicating points. This prevents NEI from being useful for discrete problems, and could also be a limitation in continuous spaces. \citet{binois17} derive conditions under which it is beneficial to replicate, and show that in some situations replication can lead to lower predictive variance across the design space than new observations. In continuous spaces, NEI will reduce uncertainty at the optimum without replicates by sampling nearby points. In our experiments this was sufficient, but incorporating a replication strategy is an area of future work \citep[see][for additional discussion on replication strategies in this setting]{jalali17}. NEI also does not give value to points outside the feasible region, due to the myopic utility function. Infeasible points may be useful for reducing model uncertainty and allowing better, feasible points in future iterations. Less myopic methods such as integrated expected conditional improvement \citep{gramacy10} measure that value. Knowledge gradient also gives value to points according to their improvement of the global model, not just their individual objective value. Incorporating utility for infeasible points into NEI could also be beneficial.

Recent work in \citet{chevalier13} and \citet{marmin16} provides an alternative to MC integration for batch Bayesian optimization using formulae for truncated multivariate normal distributions. Applying these results to the multivariate normal expectation of NEI is another promising area of future work.

For simplicity, here we assumed independence of the constraints. This could easily be replaced by a multi-task GP over the constraints for computing probability of feasibility. The sampling would then use the full covariance matrix across all constraints. The assumed independence of the objective with each constraint is required for the analytic form of the inner EI computation. Extending EI to account for correlations between objective and constraints is an open challenge.

We found that not only did NEI generally outperform PESC, but even EI+heuristics outperformed PESC in three of the four experiments. PESC has been compared to Spearmint EI on these same problems before, but in settings more similar to hyperparameter optimization than our noisy experiments setting. \citet{hernandez14} evaluated PESC on unconstrained Branin and Hartmann6 problems, but with a very low noise level: 0.03, whereas in our experiments the noise standard deviation was 5 for Branin and 0.2 for Hartmann6. \citet{hernandez15} evaluated PESC on the Gramacy problem, but with no observation noise. These previous experiments were also fully sequential, whereas ours required producing batches of 5 proposals before updating the model. \citet{shah15} evaluated predictive entropy search on unconstrained Branin and Hartmann6 problems with no noise, but with batches of size 3. They found for both of these problems that Spearmint EI outperformed predictive entropy search. \citet{metzen16} showed that entropy search can perform worse than EI because it does not take into account the correlations in the observed function values. This can cause it to be over-exploitative, and is an issue that would be exacerbated by high observation noise. The approximations required to compute and optimize PESC are sufficiently complicated that it is hard to pinpoint the source of the problem. We are interested in production optimization systems that are used and maintained by teams, and so the straightforward implementation of NEI is valuable.

Spearmint EI performed worse than EI+heuristics, despite also being an implementation of EI with heuristics. The most significant difference between the two is the way in which the constraint heuristic was implemented. EI+heuristics measured EI relative to the best point that was feasible in expectation. Spearmint EI requires the incumbent best to be feasible with probability at least 0.99 for each constraint. In our experiments with relatively noisy constraints, there were many iterations in which there were no observations with a probability of feasibility above 0.99, in which case Spearmint EI ignores the objective and proposes points that maximize the probability of feasibility. The sensitivity of the results to the way in which the heuristics are implemented provides additional motivation for ending our reliance on them with NEI. 

We demonstrated the efficacy of our method to improve the performance of machine learning infrastructure and a JIT compiler. Our method is widely applicable to many other empirical settings which naturally produce measurement error, both in online and offline contexts.

\bibliographystyle{ba}
\bibliography{noisy_ei}

\end{document}